\documentclass[sigconf, nonacm]{acmart}

\newcommand\holodoi{10.61981/ZFSH2403}
\newcommand\holopages{18-26}
\newcommand\holovolume{2}
\newcommand\holoissue{3}
\newcommand\holoyear{2024}
\newcommand\holoauthors{\authors}
\newcommand\holotitle{\shorttitle}

\newcommand\holoavailabilityurl{URL_TO_YOUR_ARTIFACTS}
\newcommand\holopagestyle{plain} 

\usepackage{balance}
\usepackage{caption}
\usepackage{subcaption}
\usepackage{graphicx}

\newcommand{\relpose}[2]{\hat{\rho}_{#1,#2}}
\newcommand{\relposegt}[2]{\rho_{#1,#2}}

\begin{document}
\setcounter{page}{17}
\title{Swazure: Swarm Measurement of Pose for Flying Light Specks}


 \author{Hamed Alimohammadzadeh}
 \email{halimoha@usc.edu}
 \orcid{0000-0003-2613-5010}
 \affiliation{%
   \institution{University of Southern California}
   \city{Los Angeles}
   \state{California}
   \country{USA}
 }

 \author{Shahram Ghandeharizadeh}
 \email{shahram@usc.edu}
 \orcid{0000-0002-1792-7879}
 \affiliation{%
   \institution{University of Southern California}
   \city{Los Angeles}
   \state{California}
   \country{USA}
 }







\begin{abstract}
One may construct a 3D multimedia display using miniature drones configured with light sources, Flying Light Specks (FLSs).  Swarms of FLSs localize to illuminate complex 3D shapes and animated sequences consistent with the coordinates of points in a point cloud.  This requires FLSs to accurately measure their pose relative to one another using sensors such as cameras. Such sensors have a sweet range in which they provide the highest accuracy.
A challenge is how an FLS tracks another FLS outside its sensor's sweet range, dictated by the point cloud data.
We address this challenge by proposing a novel technique called Swazure that solves the missing sensor data using cooperation among FLSs.
It implements {\em physical data independence} by abstracting the physical characteristics of the sensors, making point cloud data independent of the sensor hardware.  The size of an FLS relative to the minimum distance between points of a point cloud is an important parameter. With medium-sized FLSs, Swazure is able to position 100\% of the FLS's neighbors. Larger FLS sizes may result in potential obstructions that prevent Swazure from quantifying relative pose. We present two heuristics, Move Obstructing and Move Source, to address this limitation.  Our experimental results show the superiority of the Move Obstructing heuristic which resolves approximately 30\% of obstructions in the worst case scenario.

\end{abstract}

\maketitle

\pagestyle{\holopagestyle}
\begingroup\small\noindent\raggedright\textbf{Holodecks Reference Format:}\\
\holoauthors. \holotitle. Holodecks, \holovolume(\holoissue): \holopages, \holoyear.\\
\href{https://doi.org/\holodoi}{doi:\holodoi}
\endgroup
\begingroup
\renewcommand\thefootnote{}\footnote{\noindent
This work is licensed under the Creative Commons BY-NC-ND 4.0 International License. Visit \url{https://creativecommons.org/licenses/by-nc-nd/4.0/} to view a copy of this license. For any use beyond those covered by this license, obtain permission by emailing \href{mailto:info@holodecks.quest}{info@holodecks.quest}. Copyright is held by the owner/author(s). Publication rights licensed to the Holodecks Foundation. \\
\raggedright Proceedings of the Holodecks Foundation, Vol. \holovolume, No. \holoissue. \\
\href{https://doi.org/\holodoi}{doi:\holodoi} \\
}\addtocounter{footnote}{-1}\endgroup

\ifdefempty{\holoavailabilityurl}{}{
\vspace{.3cm}
\begingroup\small\noindent\raggedright\textbf{Holodecks Artifact Availability:}\\
See \url{https://github.com/flyinglightspeck/Swazure} for open source software implementations of Swazure and its data set.
\endgroup
}

\section{Introduction}
An approach to realize a 3D multimedia display~\cite{dv2023,flightpatterns2023} is to use drones configured with RGB lights, Flying Light Specks (FLSs).
Swarms of FLSs will localize~\cite{alimohammadzadeh2023swarmer,swarical2024} to illuminate complex shapes~\cite{shahram2022,integrate2023} and provide haptic interactions~\cite{shahram2021,flshaptics2023,flshaptics2024}.
These displays have diverse applications, ranging from entertainment to education and healthcare~\cite{integrate2023}.

A shape is a 3D point cloud. 
Each FLS in a 3D multimedia display is assigned a point of a point cloud indicating the position of the FLS in shape. 
FLSs are required to localize themselves to illuminate the shape. 
Each FLS lacks a line of sight in an indoor setting with a GPS satellite. Instead, it adjusts its position relative to its neighbors 
using sensors such as a camera~\cite{alimohammadzadeh2023swarmer,swarical2024}.
The neighbors of an FLS and the distance between the FLS and its neighbors are dictated by the geometry of the point cloud. To illuminate various shapes, FLSs are required to measure their relative position independent of the geometry of the point cloud.
An approach may configure each FLS with a sensor at the granularity required by the shape.
For example, if the minimum distance between the points of a point cloud is 7 millimeters and an FLS represents a point, then the sensors are required to support measuring distances as short as 7 millimeters.
This form of data dependency is too rigid for several reasons.
First, a point cloud with a shorter minimum distance (say 5 millimeters) will require either a change or an upgrade of the sensors mounted on the FLSs. 
Second, a sensor that provides 7-millimeter accuracy may either not be available or be prohibitively expensive. 

 \begin{figure}[t]
    \centering
    \captionsetup[subfigure]{justification=centering}
    \begin{subfigure}{0.32\columnwidth}
        \centering
        \includegraphics[width=0.3\textwidth]{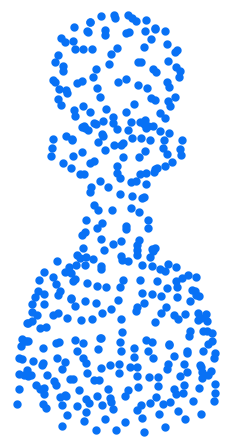}
        \caption{Chess piece, 408.}
    \end{subfigure}
    \begin{subfigure}{0.32\columnwidth}
        \centering
        \includegraphics[width=0.65\textwidth]{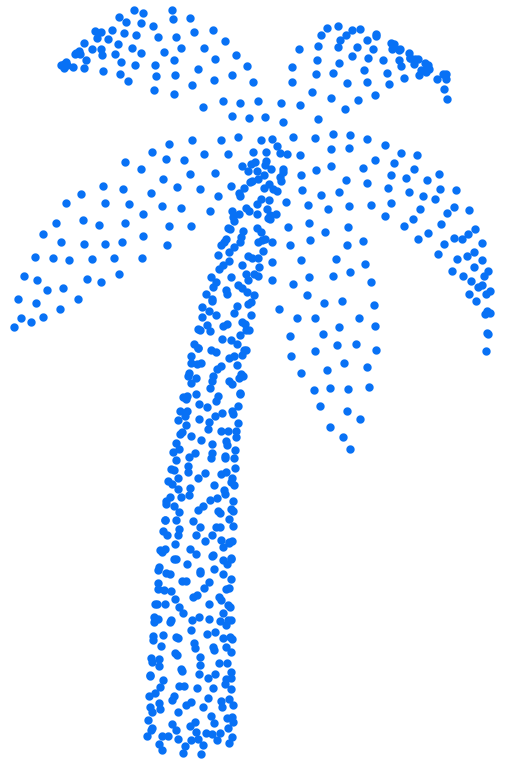}
        \caption{Palm, 725.}
    \end{subfigure}
    \begin{subfigure}{0.32\columnwidth}
        \centering
        \includegraphics[width=0.81\textwidth]{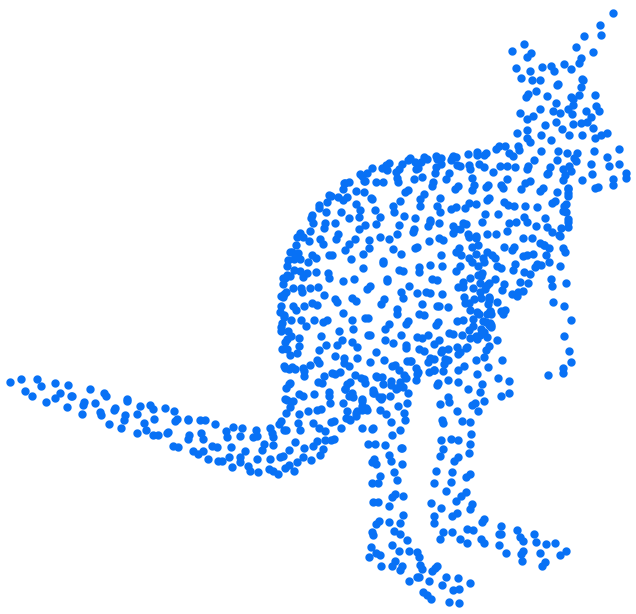}
        \caption{Kangaroo, 972.}
    \end{subfigure}
    \begin{subfigure}{0.32\columnwidth}
        \centering
        \includegraphics[width=1\textwidth]{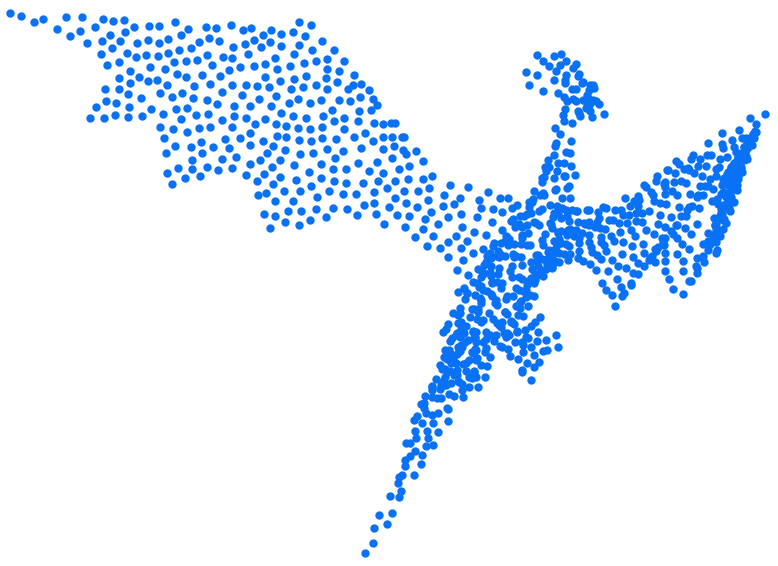}
        \caption{Dragon, 1147.}
    \end{subfigure}
    \begin{subfigure}{0.32\columnwidth}
        \centering
        \includegraphics[width=0.81\textwidth]{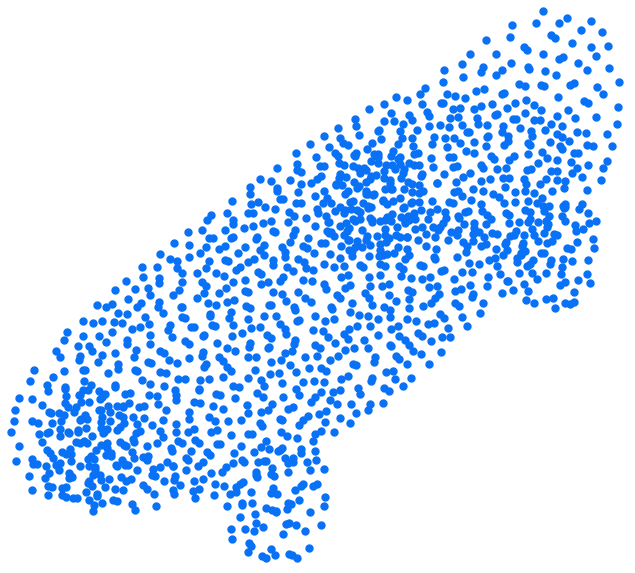}
        \caption{Skateboard, 1372.}
    \end{subfigure}
    \begin{subfigure}{0.32\columnwidth}
        \centering
        \includegraphics[width=0.88\textwidth]{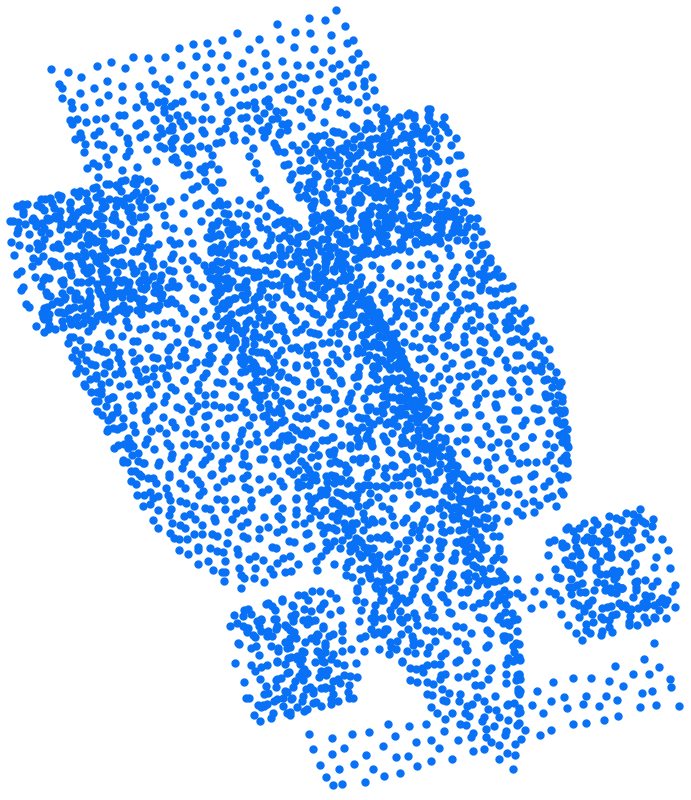}
        \caption{Racecar, 3720.}
    \end{subfigure}
    \caption{Six point clouds and their number of FLSs.} 
    \label{fig:shapes} 
\end{figure}

This study introduces the concept of {\em physical data Independence} for FLSs and 3D displays.
The concept implies that the physical hardware of FLSs is independent of the characteristics of the data (point clouds) that they illuminate.
Swarm measurement, Swazure, is an algorithm that 
implements this concept by using an abstraction of a sensor consisting of a blind, sweet, and decaying range.
Devices in the blind range are unable to quantify their relative pose to one another.
Devices in the sweet range may quantify their relative pose with high accuracy at a fixed granularity.
Devices in a decaying range can do so with some noise that impacts accuracy, granularity, or both.
This noise may be a fixed function of a physical characteristic, such as the distance between FLSs.
Swazure uses the abstraction in combination with the information exchanged by a cooperative swarm of devices to enable devices in both the blind and the decaying range to quantify their relative pose at the accuracy of the sweet range.


To illustrate, assume each FLS is configured
with a Raspberry camera and an ArUco marker to compute its pose relative to its neighbor~\cite{swarical2024}.
The camera's wide lens is unable to measure distances shorter than 5 centimeters.
This is the blind range of the wide lens.
This lens is most accurate in measuring distances between 6 to 8 centimeters at millimeter granularity.
This is the sweet range of the lens.
Beyond 8 centimeters, the error in measured distances increases as a function of the distance between the lens and the marker, the decaying range of the wide lens.
Swazure enables a swarm of FLSs to cooperate to enable those FLSs in either the blind or the decaying range 
to quantify their relative pose at the accuracy of the sweet range.
We demonstrate this with point clouds that require FLSs to measure distances of approximately 7 millimeters.

Contributions of this study include:
\begin{itemize}
    \item Swazure.  A technique that uses the communication capability of a swarm to estimate relative pose with high accuracy for distances shorter than the minimum range of the FLS sensor, its blind range. (Section~\ref{sec:swazure}.)
    \item Move Obstructing and Move Source.  Two techniques that enable Swazure to minimize the impact of FLSs obstructing the line of sight. (Section~\ref{sec:nosweet}.)
    \item An evaluation of the techniques and their tradeoffs.
    Move Obstructing is superior to Move Source. (Section~\ref{sec:eval}.) 
    \item Open source software implementations of Swazure and its data set at \url{https://github.com/flyinglightspeck/Swazure}.
\end{itemize}
The rest of this paper is organized as follows.
Section~\ref{sec:term} introduces the terminology used in this paper.
Section~\ref{sec:swazure} details Swazure.
We quantify tradeoffs associated with Swazure in Section~\ref{sec:eval}.
Section~\ref{sec:related} presents related work.
Brief conclusions are presented in Section~\ref{sec:conc}.

\section{Terminology}\label{sec:term}
We represent FLSs as spheres. Given a point cloud illuminated by FLSs, centers of FLSs are placed at the coordinate of each point of the point cloud. We determine the radius $r$ of FLSs as a ratio $\beta$ of the smallest distance $\Delta_{min}$ between a pair of points, $r=\beta \times \Delta_{min}$.

We represent the estimated relative pose of FLS $f_1$ to FLS $f_n$ using a position vector $\relpose{1}{n}$ where $f_n$ is the vector's head and $f_1$ at the origin is its tail.
Note that $\relpose{1}{n} = -\relpose{n}{1}$. To distinguish between estimated and ground truth vectors, we represent the relative pose of corresponding FLSs in the ground truth by $\relposegt{1}{n}$. 
This vector defines $(\Delta, \theta, \phi)$ in the spherical coordinate system.
$\Delta$ is the distance between $f_1$ and $f_n$.
It is defined by the magnitude of $\relpose{1}{n}$, $\Delta=||\relpose{1}{n}||_2=\sqrt{x^2+y^2+z^2}$.
$\theta$, \textit{inclination}, is the angle between the vector and the z-axis. 
$\phi$, \textit{azimuth}, is the angle between the projection of the vector on the x-y plane and the x-axis.

We assume the sensor used by a source FLS $f_1$ to measure its pose relative to a target FLS $f_n$ has a fixed range $R$ beyond which it cannot measure its pose.
It is possible for $R$ to be unbounded, i.e., $R=\infty$.
Swazure divides $R$ into three segments:
blind, sweet, and decaying.
We define these in turn.

When $f_1$ is in the blind range of $f_n$, $f_1$ may not measure its distance relative to $f_n$ at all.  
$f_1$ is termed a blind neighbor of $f_n$.
It may have many such neighbors.
The characteristics of a point cloud dictate the exact number of neighbors for $f_1$.

When the distance separating $f_1$ from $f_n$ is in the sweet range of $f_1$'s sensor, $f_1$ may measure its pose relative to $f_n$ with a high accuracy.
$f_n$ is termed a sweet neighbor of $f_1$.

When the distance between $f_1$ and $f_n$ is in the decaying range of $f_1$'s sensor, $f_1$ is able to measure its pose relative to $f_n$ with an error that is a function of a physical property between $f_1$ and $f_n$.
$f_n$ is termed a decaying neighbor of $f_1$.
In this paper, we assume distance as the physical property.

\begin{figure}[t]
    \centering
    \begin{subfigure}{0.45\columnwidth}
        \centering
        \includegraphics[width=\textwidth]{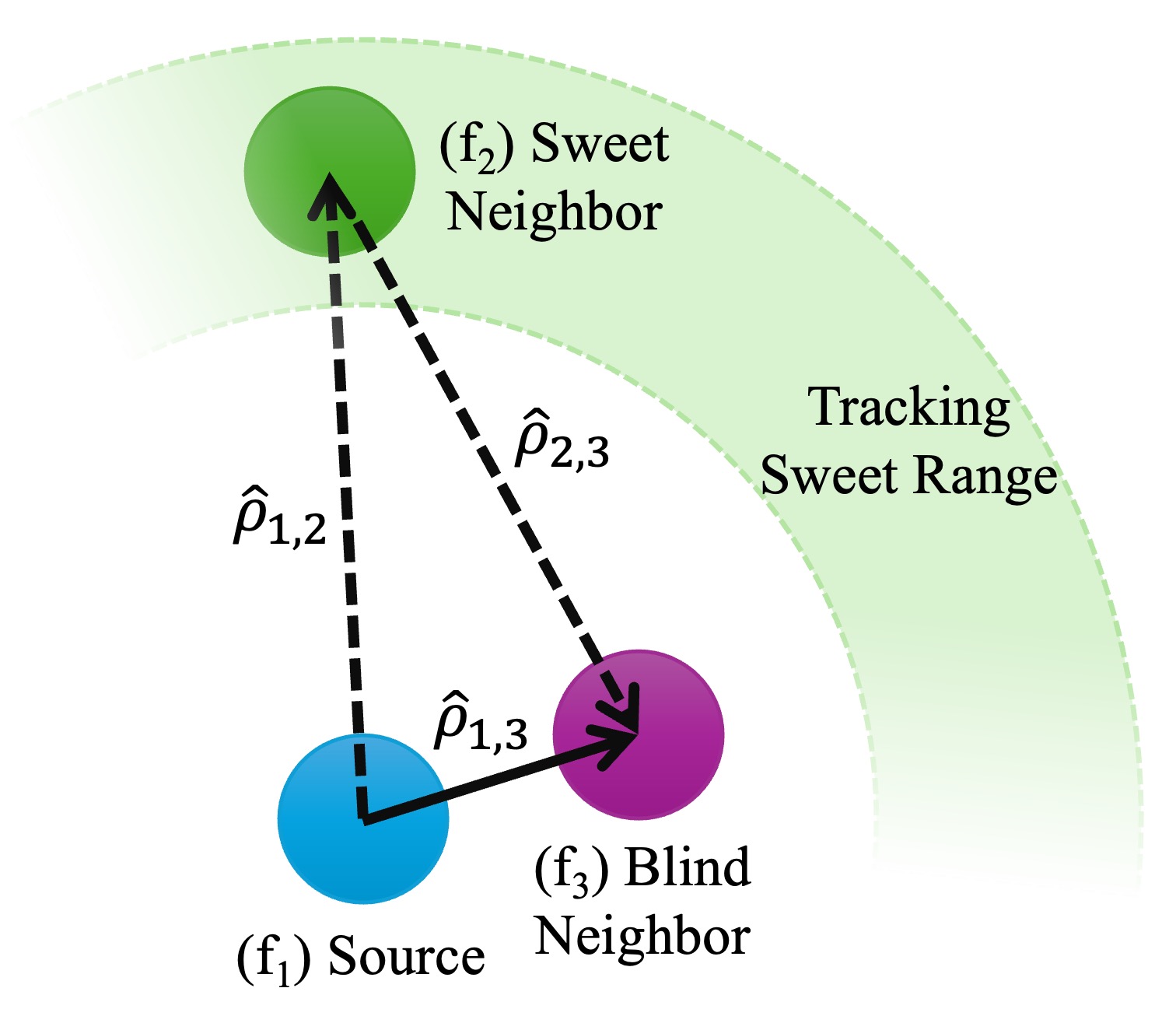}
        \caption{
        }
        \label{fig:technique}
    \end{subfigure}
    \begin{subfigure}{0.45\columnwidth}
        \centering
        \includegraphics[width=\textwidth]{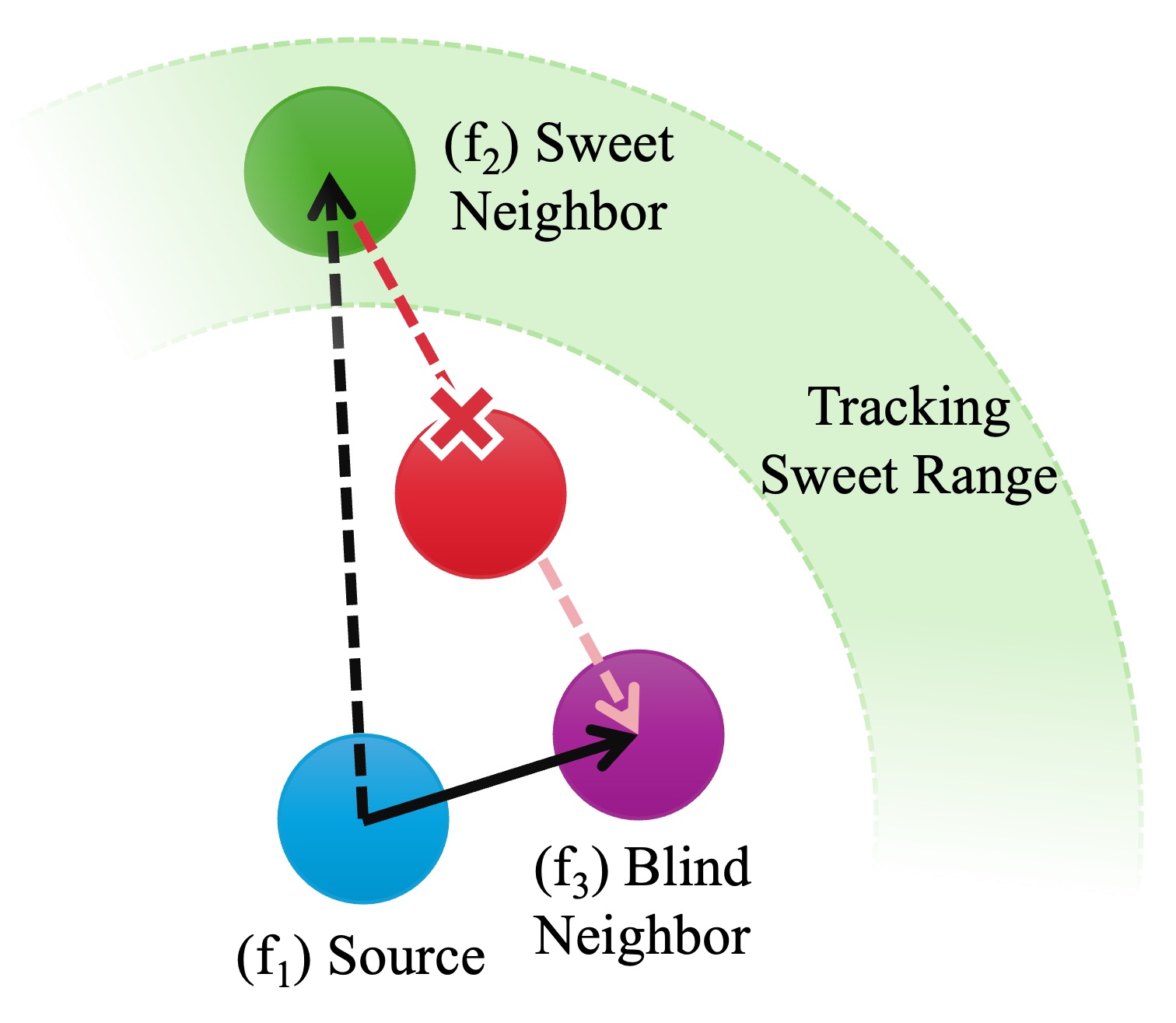}
        \caption{
        }
        \label{fig:obstruct}
    \end{subfigure}
    \caption{(a) 
    Source FLS uses a sweet neighbor to compute its pose relative to a blind neighbor, $\relpose{1}{3}=\relpose{1}{2}+\relpose{2}{3}$.
    (b) An obstructing FLS blocks the line of sight between the sweet and blind neighbor.}
\end{figure}

We assume the technique used to identify sweet neighbors is different than the technique used to quantify the relative pose. 
Hence, it is possible for a blind FLS pair, $f_1$ and $f_n$, to share a sweet neighbor with one FLS (either $f_1$ or $f_n$) lacking line of sight to the sweet neighbor.
Sweet neighbors may be detected based on FLSs exchanging messages with one another.
The sensor used to quantify orientation may be a camera. 
However, $f_1$'s camera may lack line of sight with the sweet neighbor.
One reason may be a different FLS is in between $f_1$ and its sweet neighbor,  obstructing $f_1$'s line of sight to $f_n$'s ArUco marker.
Figure~\ref{fig:obstruct} illustrates this scenario with $f_1$ (blue) and $f_3$ (purple) FLSs using the messages sent by $f_2$ (green) FLS to identify it as a sweet neighbor.
However, the red FLS blocks the line of sight of the purple FLS, preventing it from quantifying its pose relative to the green FLS.


\section{Swazure}\label{sec:swazure}
Swazure employs FLSs in the sweet range of one another to compute the relative pose of a source FLS, $f_1$, to a target FLS, $f_n$.
If $f_n$ is a sweet neighbor of $f_1$, then $f_1$ computes its pose using its tracking sensor directly. 
Otherwise, it identifies the intermediate FLSs within its sweet range and communicates with them to compute its relative pose to $f_n$ as follows.
First, it computes $\Omega_1=(f_1, f_2, ..., f_n)$ as a sequence of FLSs that defines a unique sweet path from the source ($f_1$) to the target ($f_n$).
In other words, the distance between $f_i$ and $f_{i+1}$ is in the sweet range of $f_i$'s tracking device. 
The relative pose between the source and target FLSs is computed by adding\footnote{Simply add the x,y,z coordinates of the vectors.} the relative poses along the edges of this path, $\relpose{1}{n}=\sum_{k=1}^{|\Omega_1|-1}{\relpose{k}{k+1}}$.

There may be $\lambda$ unique paths connecting $f_1$ to $f_n$.
Each is identified by a unique $\Omega_i$, $1 \leq i \leq \lambda$.
$\Omega_i$ consists of $|\Omega_i|$ FLSs.
We concpeturalize it as a graph with $|\Omega_i|$ vertices and $|\Omega_i|-1$ edges.
The edges are from $f_1$ to $f_2$, $f_2$ to $f_3$, $f_{|\Omega_i|-1}$ to $f_{|\Omega_i|}$.
Each edge has a length, $length(f_i, f_{i+1})$, denoting the distance between two FLSs in a hop.
Each $\Omega_i$ consists of $|\Omega_i|-1$ hops and has a total length $l_i=\sum_{j=1}^{|\Omega_i|-1}length(f_j,f_{j+1})$.

With $\lambda$ candidate paths, 
Swazure may select the shortest path to estimate the pose between two blind FLSs.
Shortest is defined as either the fewest number of hops (Fewest-Hops) or the shortest length (Shortest-Length).



To illustrate, consider a path $\Omega_1$ consists of two hops, $\Omega_1=(f_1, f_2, f_3)$. 
See Figure \ref{fig:technique}.
$f_1$ computes its relative pose to its blind neighbor $f_3$, $\relpose{1}{3}$, by
summing its relative pose to its sweet neighbor $f_2$ ($\relpose{1}{2}$) with $f_2$'s relative pose to its sweet neighbor $f_3$ ($\relpose{2}{3}$), i.e., $\relpose{1}{3}$=$\relpose{1}{2}+\relpose{2}{3}$.
This path consists of two hops, $f_1$ to $f_2$ and $f_2$ to $f_3$.
Each hop has a fixed length defined by the magnitude of its corresponding vector, e.g., length($f_1$,$f_2$)=$||\relpose{1}{2}||_2$.
The total length $l_1$ of the path is the sum of the length of $f_1$ to $f_2$ and the length of $f_2$ to $f_3$.

\begin{figure*}[t]
    \centering
    \begin{subfigure}{0.6\columnwidth}
        \centering
    \centering
    \includegraphics[width=\columnwidth]{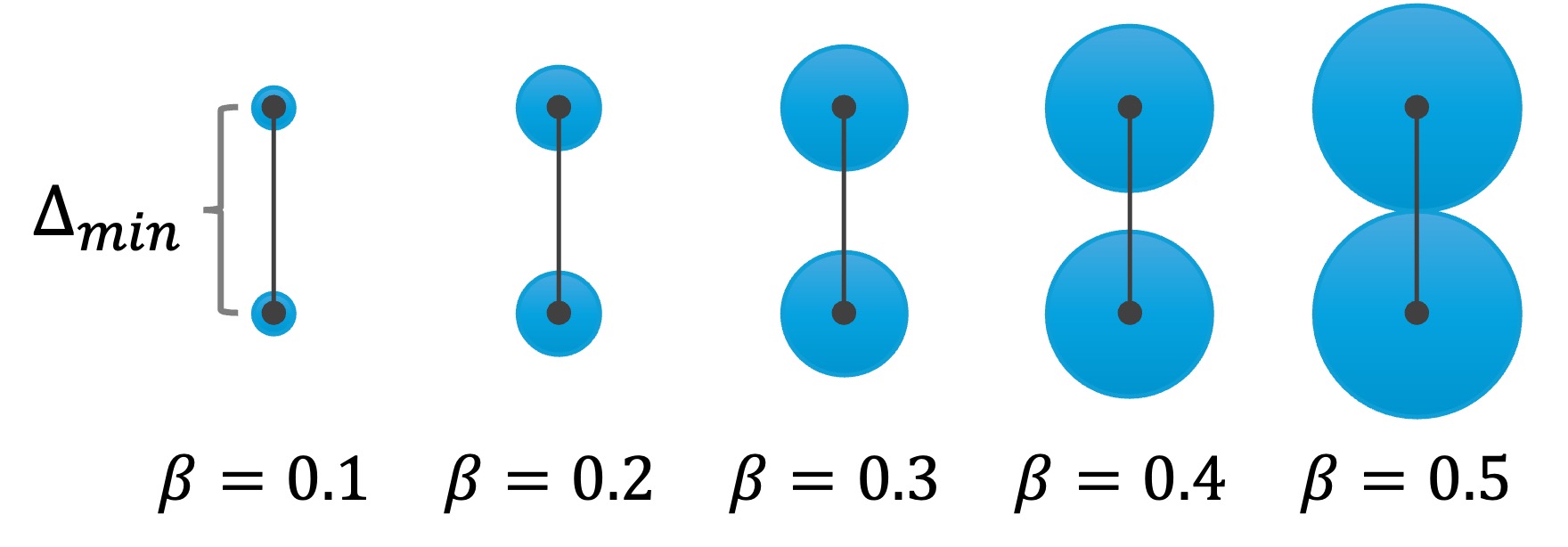}
    \caption{Different $\beta$ values.
    }
    \label{fig:showbeta}
    \end{subfigure}
    \hfill
    \begin{subfigure}{0.6\columnwidth}
        \centering
        \includegraphics[width=\columnwidth]{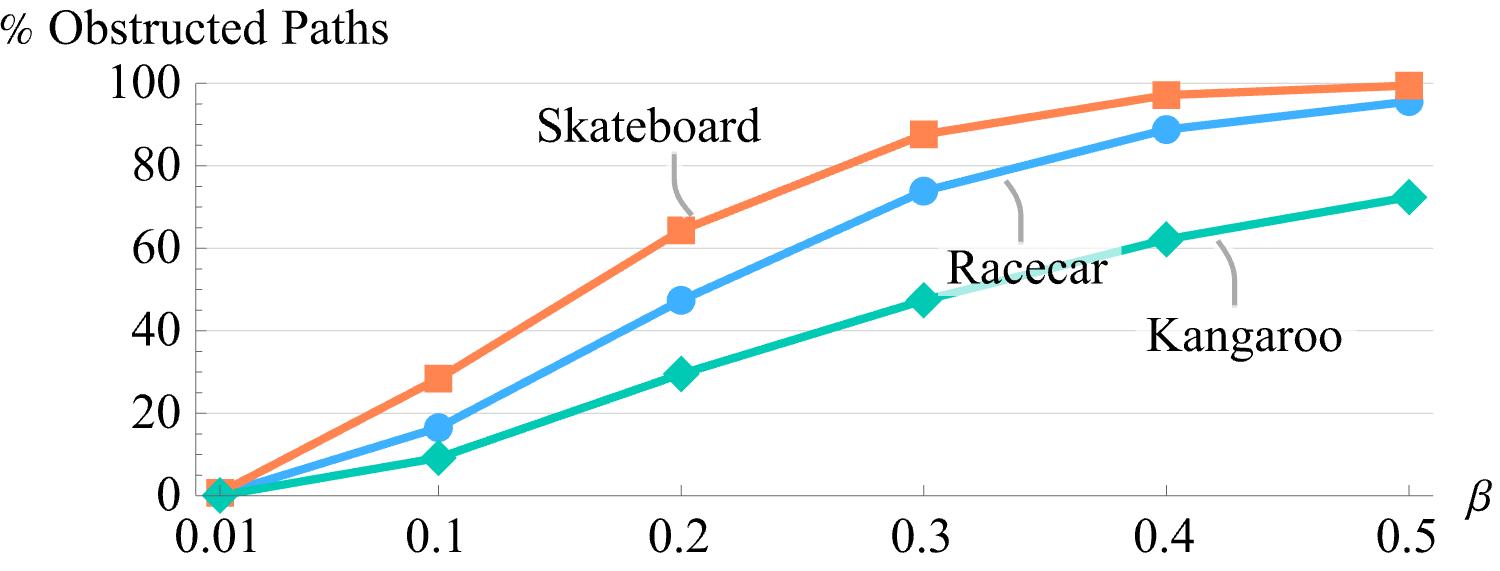}
        \caption{Percentage of obstructing paths.}
        \label{fig:obstructed-path}
    \end{subfigure}
    \hfill
    \begin{subfigure}{0.6\columnwidth}
        \centering
        \includegraphics[width=\columnwidth]{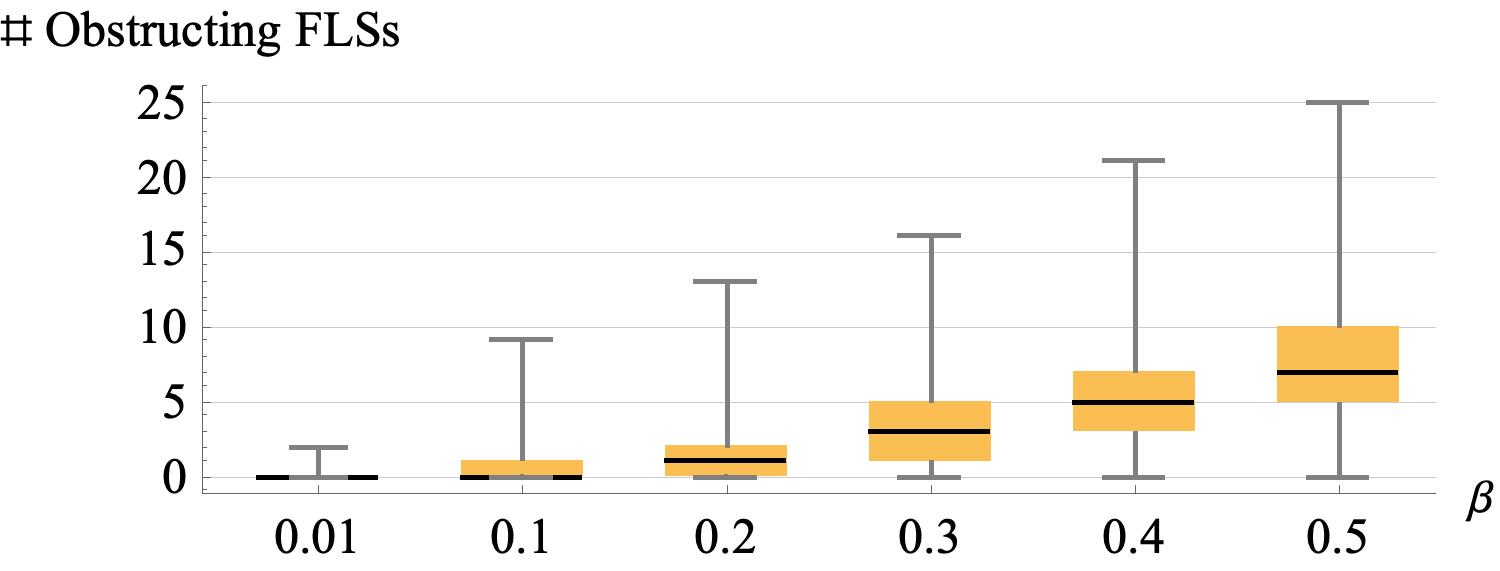}
        \caption{Number of FLSs in an obstruction.}
        \label{fig:obstructing-fls-box}
    \end{subfigure}
    \caption{(a) $\beta$ is the ratio of FLS radius to the minimum distance $\Delta_{min}$ between FLSs, it impacts (b) the percentage of obstructed paths and (c) the number of FLSs participating in obstruction.}
\end{figure*}
\subsection{No Sweet Paths}\label{sec:nosweet}
The geometry of a point cloud dictates the number of sweet paths between two blind FLSs.
Each path may have one or many obstructing FLSs.  
$\beta$ is an important parameter, see Figure~\ref{fig:showbeta}.
It is the ratio of the radius of an FLS to the minimum distance between them, $\Delta_{min}$.
It dictates both the number of sweet paths with an obstructing FLS and the number of obstructing FLSs for that path.
To illustrate,
Figure~\ref{fig:obstructed-path} shows the percentage of sweet paths with an obstructed FLS for three different shapes as a function of $\beta$.
It highlights the geometry of a shape as a significant factor.
While almost all sweet paths are obstructed with the Skateboard and $\beta$=0.5, approximately 60\% are obstructed with the Kangaroo and the same\footnote{With $\beta=0.5$, the number of FLSs that are tightly packed next to one another equals the number of shortest distance paths between FLSs.
This number is 9 with the Chess piece, 12 with the Dragon, 18 with the Kangaroo, 15 with the Palm, 57 with the Race Car, and 22 with the Skateboard.} $\beta$ value.
There is also a significant variation in the number of obstructing FLSs per sweet path.
The box plot for the Skateboard highlights this variation. See Figure~\ref{fig:obstructing-fls-box}.


\begin{figure}
    \centering
    \includegraphics[width=.75\columnwidth]{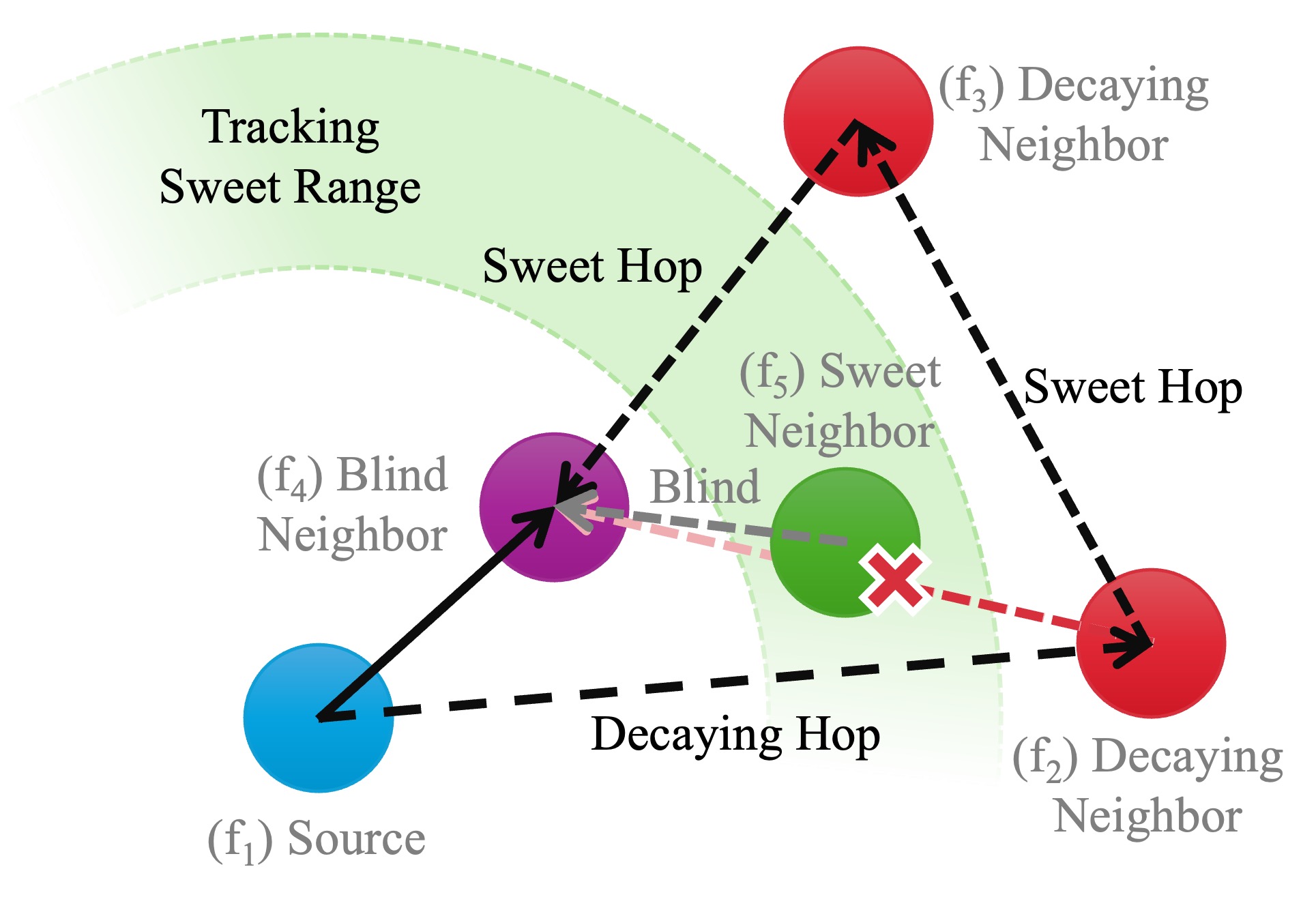}
    \caption{
    Source FLS $f_1$ uses a decaying hop followed by two sweet hops to compute its pose relative to its blind neighbor $f_4$.
    $\relpose{1}{4}=\relpose{1}{2}+\relpose{2}{3}+\relpose{3}{4}$.}
    \label{fig:multihop}
\end{figure}

In the absence of sweet paths, Swazure may use decaying paths.
In other words, the path between the source and target FLS may use neighboring FLSs in the decaying range.
Figure \ref{fig:multihop} shows an example with $f_1$ (blue) FLS wanting to estimate its relative pose to its blind neighbor $f_4$ (purple).
$f_5$ (green) FLS is a sweet neighbor of $f_1$.
However, $f_1$ may not use this neighbor because it is a blind neighbor of $f_4$.
Hence, $f_1$ uses $f_2$ with a decaying hop.
The subsequent hops from $f_2$ to $f_3$ and $f_3$ to $f_4$ are sweet.
Note that $f_2$ is a sweet neighbor of $f_4$.
However, $f_5$ (green) is blocking its line of sight with $f_4$.
This technique results in a high error in the computed relative pose as the error increases as a function of the distance between FLSs.
Note that with this technique, Swazure may use a combination of sweet and decaying hops.

It is possible that there are no paths between two blind neighbors.
This is especially true with high values of $\beta$, $\beta \geq$0.4.
The first two columns of Table~\ref{tab:nopath} show the percentage of blind neighbors with no paths for different shapes.
While it is zero with the Palm and the Dragon, it is as high as 31\% with the Chess piece and $\beta$=0.5.  
(The percentage of blind neighbors with no path is zero with $\beta \leq$0.3.)
The last three columns of Table~\ref{tab:nopath} show the percentage of decaying paths employed by Swazure with different shapes.
They highlight the significance of the geometry of a shape and $\beta$. 

\begin{table}[h]
    \centering
    \caption{\% blind neighbors with no/decaying paths.}
    \begin{tabular}{|l|c c|c c c|}
    \hline
    \hline
             & 
        \multicolumn{2}{c|}{No Paths} & \multicolumn{3}{c|}{Decaying Paths} \\
    \cline{2-6}
            & $\beta$=0.4    & $\beta$=0.5 & $\beta$=0.3 & $\beta$=0.4    & $\beta$=0.5  \\
    \hline
Chess Piece & 2.96\%  & 30.88\% & 0 & 1.27\% & 16.77\% \\
Palm       & 0      & 0    & 0 & 1.15\% & 11.3\% \\
Kangroo     & 0.09\% & 0.38\% & 0.38\% & 0.97\% & 2.08\% \\
Dragon      & 0      & 0     & 0.43\% & 9.15\% & 31.01\% \\
Skateboard  & 0.36\%  & 17.92\% & 0     &   0    & 12.42\% \\
Race Car     & 0  & 0.03\% & 0 & 0 & 1.01\% \\
    \hline
    \hline
    \end{tabular}
    \label{tab:nopath}
\end{table}

\begin{figure}
    \centering
    \begin{subfigure}{0.4\columnwidth}
        \includegraphics[width=\textwidth]{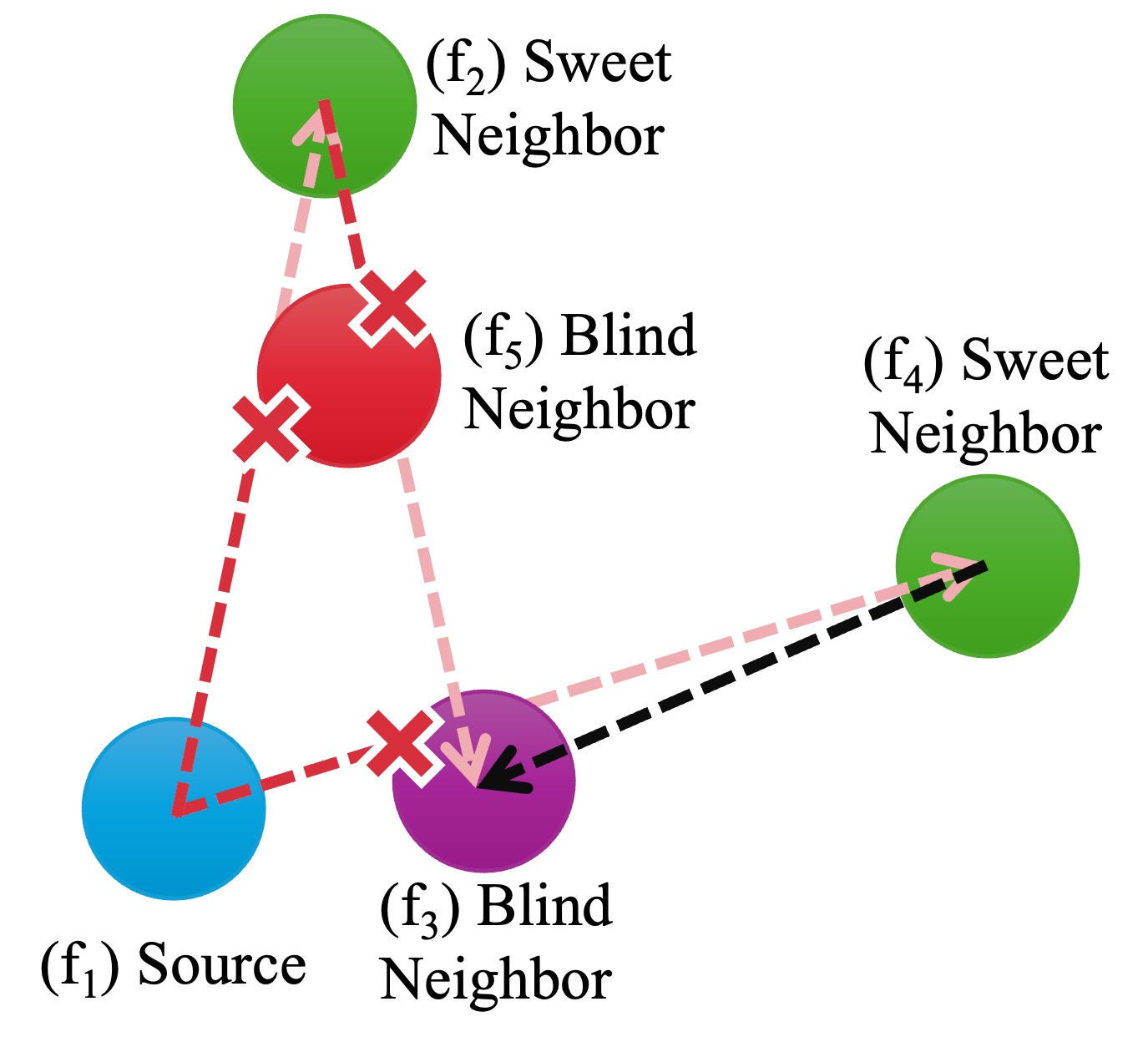}
        \caption{Before.}
        \vspace{1.5cm}
    \end{subfigure}
    \hspace{.5cm}
    \begin{subfigure}{0.4\columnwidth}
        \centering
        \begin{subfigure}{\textwidth}
            \centering
            \includegraphics[width=\textwidth]{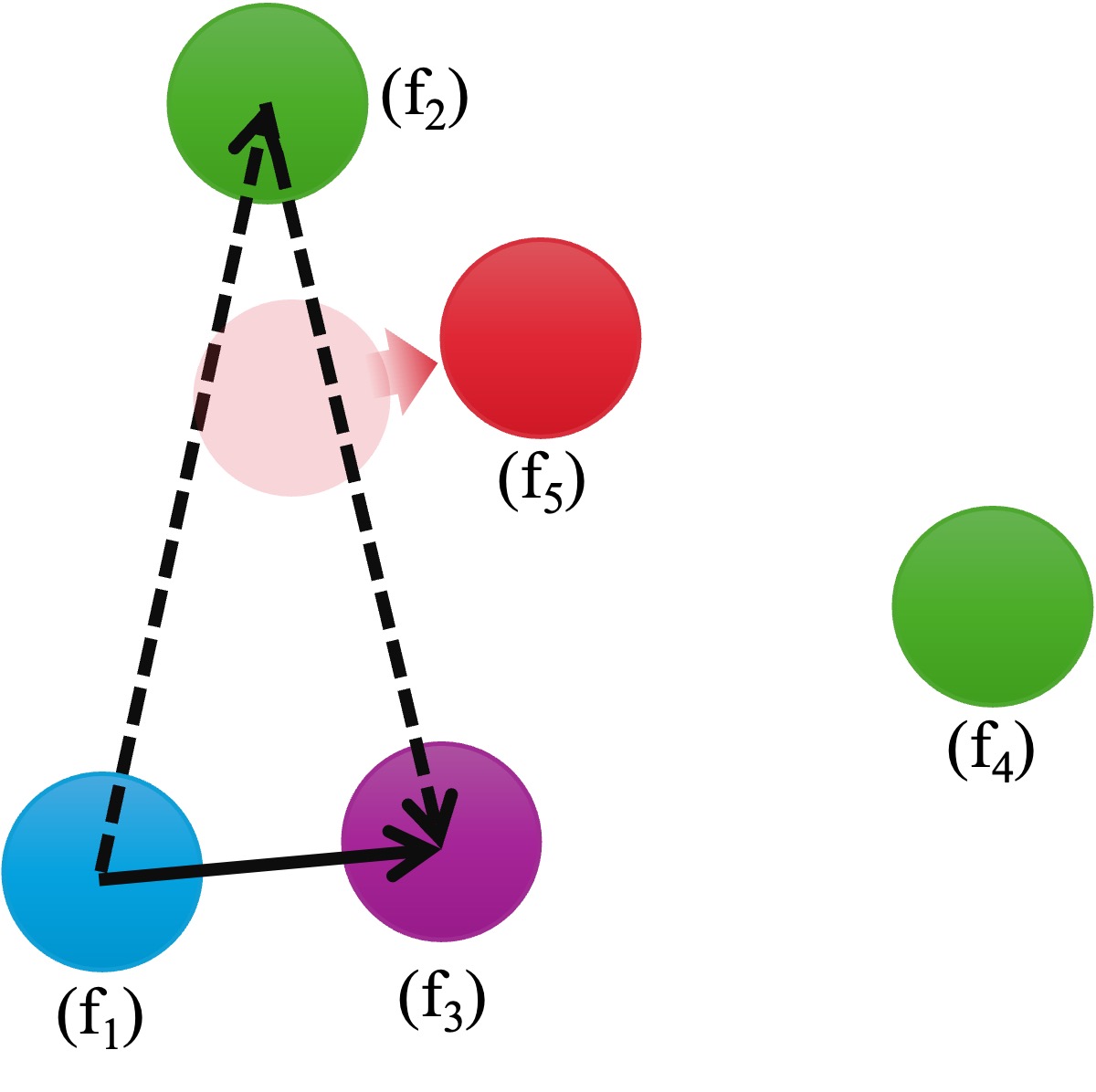}
            \caption{Move Obstructing.}
        \end{subfigure}
        \begin{subfigure}{\textwidth}
            \centering
            \includegraphics[width=\textwidth]{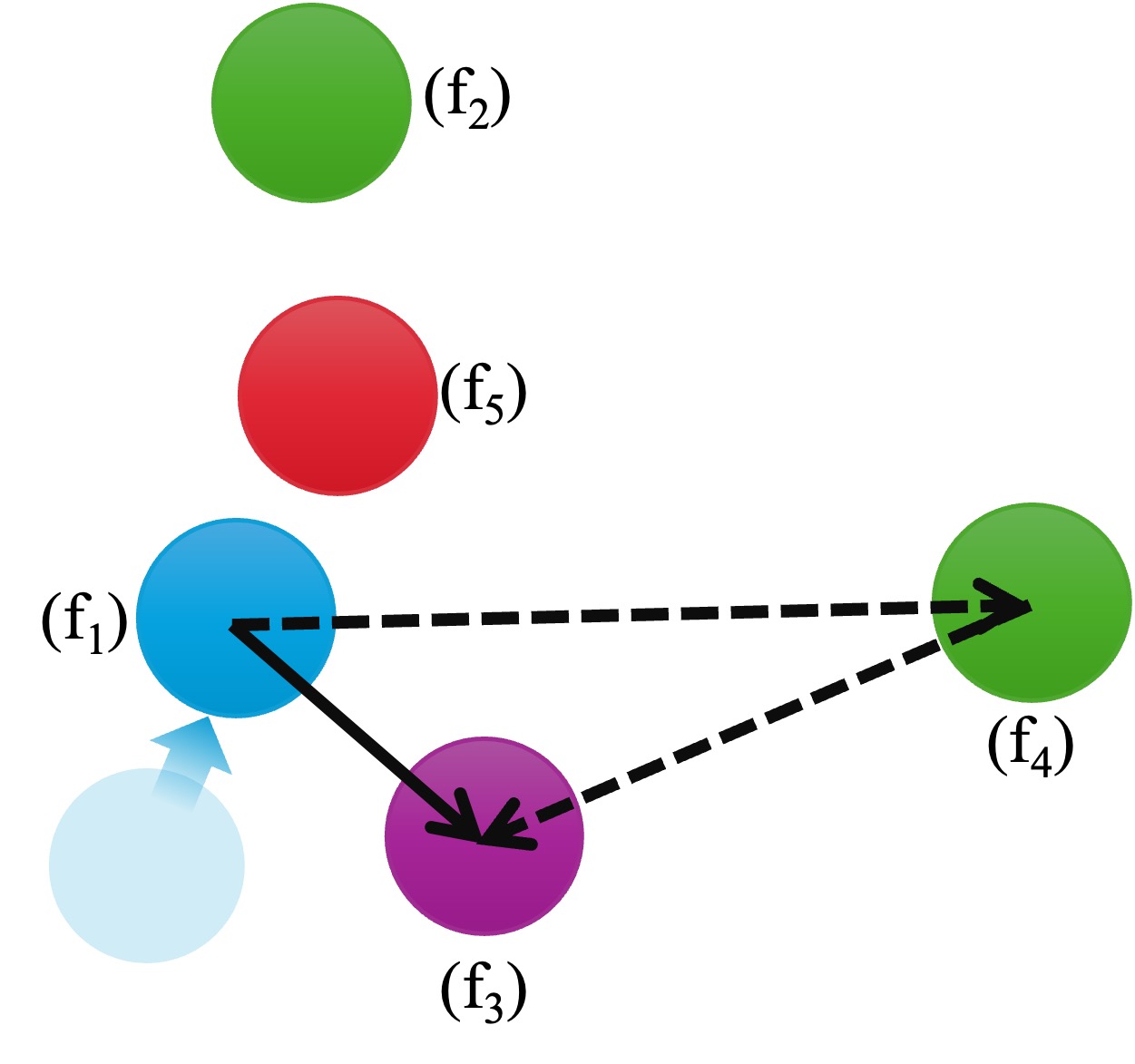}
            \caption{Move Source.}
        \end{subfigure}
    \end{subfigure}
    \caption{Move Obstructing and Move Source.}
    \label{fig:move-obs-source}
\end{figure}

Below, we describe two heuristics to reduce the number of blind neighbors with no paths and replace decaying paths with sweet ones.

\subsection{Move Obstructing}\label{sec:moveobstructing}
A sweet path can be created between a blind pair by moving the FLSs that obstruct the line of sight between FLSs and their sweet neighbors. In this technique, FLSs detect obstructing FLSs and command them to move along a vector. Each obstructing FLS moves the minimum distance required to resolve the line of sight in a direction perpendicular to the line connecting the desired FLSs. 



\begin{figure}
    \centering
    \begin{subfigure}{0.4\columnwidth}
        \includegraphics[width=\textwidth]{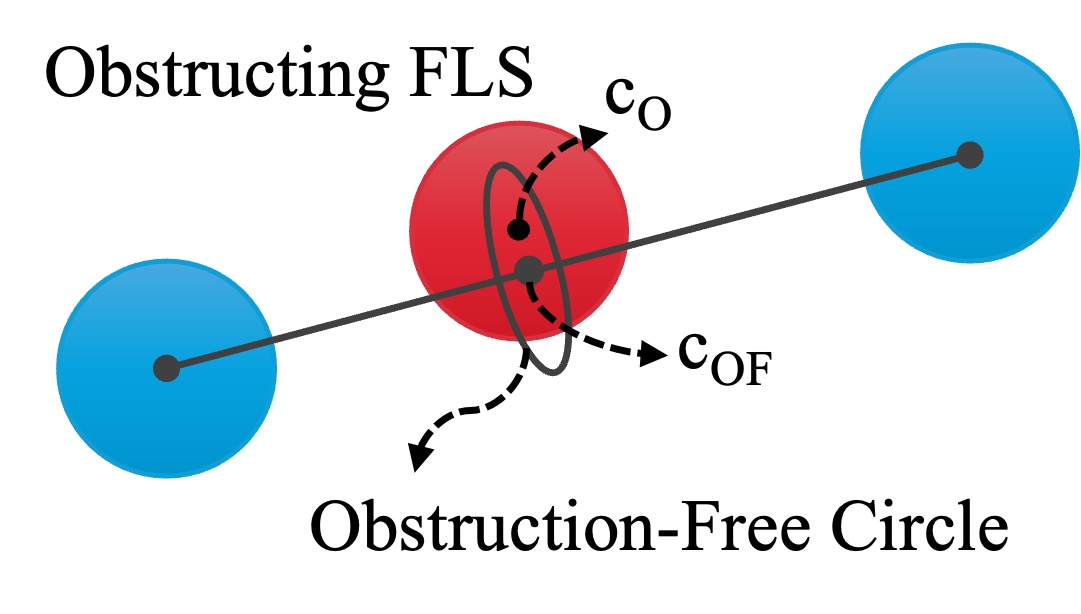}
        \caption{Obstructing FLS.}\label{fig:defObst}
        \vspace{1cm}
    \end{subfigure}
    \hspace{.5cm}
    \begin{subfigure}{0.4\columnwidth}
        \centering
        \begin{subfigure}{\textwidth}
            \centering
            \includegraphics[width=\textwidth]{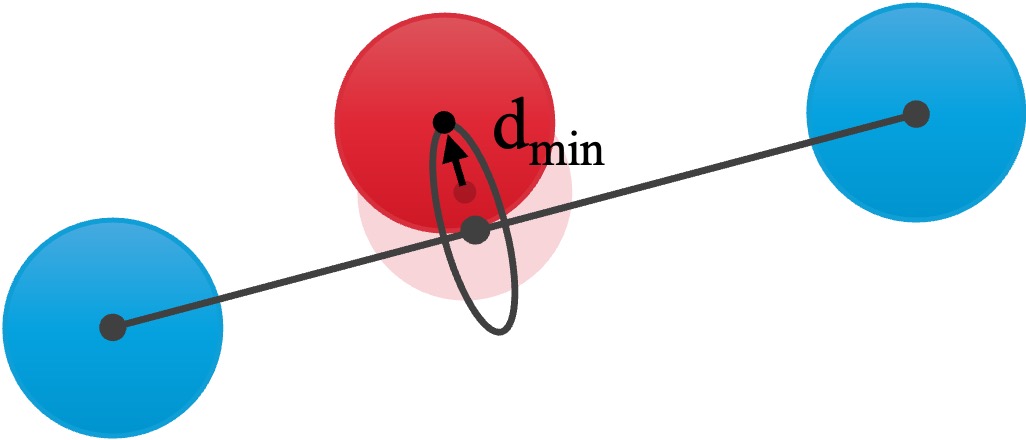}
            \caption{Move along $d_{min}$.}
        \end{subfigure}
        \begin{subfigure}{\textwidth}
            \centering
            \includegraphics[width=\textwidth]{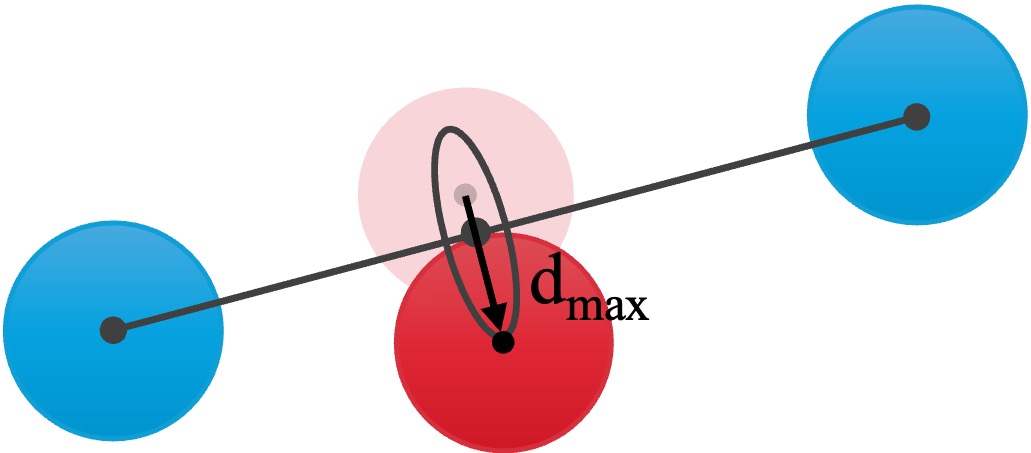}
            \caption{Move along $d_{max}$.}
        \end{subfigure}
    \end{subfigure}
    \caption{The minimum and maximum distance an obstructing FLS has to move to resolve the obstruction using the Move Obstructing heuristic.}
    \label{fig:move-obs-source}
\end{figure}

More formally, we define the line of sight as a line from the center of a source FLS $f_1$ to its sweet neighbor.
An obstructing FLS intersects this line; see Figure~\ref{fig:defObst}.
We compute an obstruction-free circle around the obstructing FLS.
The center of this circle is the closest point on the line to the center of the obstructing FLS.
The radius of this circle is equal to the radius of the FLS.
We identify the collision-free portions of this circle.
If none exists, then this technique fails.
Otherwise, we move the obstructing FLS to a collision-free portion that is closest to its current location.
To identify a collision-free portion, we consider a collision sphere for each of the obstructing FLS's neighbors. 
The radius of these spheres is twice that of the FLSs. We compute the intersection of these spheres with the obstruction-free circle, identifying them as potential collisions.  We remove this portion of the circle. Its remainder is the collision-free portion.

The minimum and maximum vector for moving the FLS are defined as $d_{min}=(\frac{r}{|c_O - c_{OF}|}-1)(c_O - c_{OF})$ and $d_{max}=(\frac{r}{|c_O - c_{OF}|}+1)(c_O - c_{OF})$.
The magnitude of these vectors bound the minimum and maximum distance moved by the obstructing FLS.
See Figure~\ref{fig:move-obs-source}.


Table~\ref{tab:nopathMO} shows 
This technique is effective in converting no paths to sweet paths.
The values in the first two columns of Table~\ref{tab:nopathMO} are lower than those in Table~\ref{tab:nopath}.
It is also effective in converting decaying paths into sweet paths, resulting in lower percentages in 
the last three columns of Table~\ref{tab:nopathMO} when compared with Table~\ref{tab:nopath}.

\begin{table}[h]
    \centering
    \caption{\% blind neighbors with no/decaying paths using Move Obstructing.}
    \begin{tabular}{|l|c c|c c c|}
    \hline
    \hline
             & 
        \multicolumn{2}{c|}{No Paths} & \multicolumn{3}{c|}{Decaying Paths} \\
    \cline{2-6}
            & $\beta$=0.4    & $\beta$=0.5 & $\beta$=0.3 & $\beta$=0.4    & $\beta$=0.5  \\
    \hline
Chess Piece & 0.07\% & 21.43\% & 0 & 0.02\% & 11.65\% \\
Palm       & 0      & 0    & 0 &  0.13\% & 6.3\% \\
Kangroo     & 0.01\% & 0.26\% &  0 & 0.02\% & 1.17\% \\
Dragon      & 0      & 0     &  0 & 0.51\% & 23.34\% \\
Skateboard  & 0.02\%  & 12.85\% & 0     &   0    & 8.79\% \\
Race Car     & 0  & 0.02\% & 0 & 0 & 0.74\% \\
    \hline
    \hline
    \end{tabular}
    \label{tab:nopathMO}
\end{table}

\subsection{Move Source}\label{sec:movesource}
This technique finds a common sweet neighbor and moves the source to establish a path. The source FLS starts to explore the area in its vicinity until it finds at least one common sweet neighbor with the target FLS. 
It may explore a prespecified set of directions. 
If it encounters a collision in a certain direction, it will stop exploring that direction.  
A threshold bounds the distance explored in each direction. After finding a common sweet neighbor, the source FLS uses it to compute its pose relative to the target FLS.

Our specific implementation is as follows.
The source FLS contacts the target FLS for its sweet neighbors.
It identifies how many of these neighbors are in its line of sight when it adjusts its position by a vector with the magnitude=$\delta$R where R is the radius of its sphere.
It considers 26 directions using itself as the center of a logical sphere.
These include movement along the 3 primary axes, along the diagonals in the planes defined by these axes, and along the space diagonals that span across all three axes.
It starts with $\delta$=1 and increments it by 1.
A direction is dropped from consideration when it results in a collision.
It terminates either as soon as it identifies a sweet path to its target or some threshold T on the value of $\delta$ is reached.


\begin{table}[h]
    \centering
    \caption{\% blind neighbors with no/decaying paths using Move Source with T=10.}
    \begin{tabular}{|l|c c|c c c|}
    \hline
    \hline
             & 
        \multicolumn{2}{c|}{No Paths} & \multicolumn{3}{c|}{Decaying Paths} \\
    \cline{2-6}
            & $\beta$=0.4    & $\beta$=0.5 & $\beta$=0.3 & $\beta$=0.4    & $\beta$=0.5  \\
    \hline
Chess Piece &  1.55\% & 20.42\% & 0 & 0.68\% & 12.83\% \\
Palm       & 0      & 0    & 0 &  0.65\% & 7.58\% \\
Kangroo     &  0.08\% & 0.37\% & 0.26\% & 0.78\% & 1.67\% \\
Dragon      & 0      & 0     &  0.24 & 5.63\% & 22.44\%  \\
Skateboard  & 0.18\%  & 13.52\% & 0     &   0    & 9.22\% \\
Race Car     & 0  & 0.01\% & 0 & 0 & 0.7\% \\
    \hline
    \hline
    \end{tabular}
    \label{tab:nopathMS10}
\end{table}

We considered different threshold values ranging from 2 to 10.
Table~\ref{tab:nopathMS10} shows the results with T=10.
A higher value of T enables a source FLS to search a larger space around itself.
This increases the chances of it finding a sweet path that terminates its search.
Move Source reduces the percentage of blind neighbors with no paths.
Compare the first two columns of Table~\ref{tab:nopath} with those of Table~\ref{tab:nopathMS10}.
It also reduces the number of decaying paths by converting them to sweet ones.
Compare the last three columns of Table~\ref{tab:nopath} with those of Table~\ref{tab:nopathMS10}.

\subsection{A Comparison}
Tables~\ref{tab:nopathMO}-\ref{tab:nopathMS10} show Move Obstructing is superior to Move Source.
It reduces the percentage of blind neighbors with no paths.
In addition, it replaces a higher percentage of decaying paths with sweet ones.
Table~\ref{tab:MOMSdist} shows Move Obstructing results in a significantly lower distance moved by FLSs.
When considering the maximum distance moved, the difference is orders of magnitude lower with Move Obstructing.

\begin{table}[h]
    \centering
    \caption{Average (max) distance (cm) of FLSs' movement, $\beta$=0.5.}\label{tab:MOMSdist}
    \begin{tabular}{|l|c|c c|}
    \hline
    \hline
             & 
        Move  & \multicolumn{2}{c|}{Move Source} \\
    \cline{3-4}
            &  Obstructing  & T=2 & T=10   \\
    \hline
Chess Piece &  0.34 (1.51) & 23.7 (45.4) & 150.7 (609) \\
Palm        &  0.54 (2.34) & 25.1 (42.7) & 155 (574.4) \\
Kangroo     &  0.62 (2.5) & 25.7 (44.14) & 215.3 (745.2) \\
Dragon      &  0.55 (2.72) & 23.8 (46.3) & 176.8 (744.5) \\
Skateboard  &  0.43 (2.26) & 22.1 (43.6) & 109.7 (622) \\
Race Car     &  0.3 (1.49)  & 20.9 (45.3) & 88.5 (709.8) \\
    \hline
    \hline
    \end{tabular}
    
\end{table}

\subsection{Motion Estimation: Velocity and Acceleration}

An FLS can estimate the velocity and/or acceleration of another FLS using a sequence of position observations. It can estimate its velocity by computing the derivative of position with respect to time. Using finite differences we have: $v(t_i) \approx \frac{p(t_{i+1}) - p(t_{i-1})}{2 \Delta t}$ where $\Delta$ t is the time interval between measurements. Similarly, acceleration is $a(t_i) \approx \frac{v(t_{i+1}) - v(t_{i-1})}{2 \Delta t}$.

In Swazure, when an FLS ($f_1$) wants to estimate the velocity of another FLS ($f_3$) to which it does not have a line of sight, it communicates with an alternative FLS ($f_2$) to receive position measurements. $f_2$ computes the position of $f_3$ and transmits the sequence of position observations to $f_1$. There are two types of delay in this process: the delay in computing the position based on sensory data and the delay in communication. To compensate for these delays, the receiver FLS can maintain a sequence of recent samples and predict the current velocity or acceleration using extrapolation and filtering techniques such as the Kalman filter \cite{761918}.



\section{Evaluation}\label{sec:eval}
We simulated Swazure using FLSs configured with ArUco markers and a Raspberry camera~\cite{swarical2024}.
The camera is small, lightweight, and ready for use with a drone.
We use its wide lens with a minimum focus range of 5 cm.
Smaller distances are in its blind range.
Its sweet range is 6-8 cm.
Beyond 8 cm, its percentage error in pose increases as a function of the distance between the wide lens and the ArUco marker~\cite{swarical2024}.
We model this error using the following quadratic function:
$f(x) = 0.00180987x^2 -0.02756392x +  0.11561755$.
The simulator abstracts FLSs as spheres with radius $r$.
It assumes a camera and an identifying ArUco marker are placed on the sphere's surface, enabling an FLS to quantify its relative pose to another FLS.

The radius $r$ determines the amount of obstruction between FLSs.
With $\beta > 0.5$, a pair of FLSs will collide with one another. 
The simulator removes the colliding FLSs prior to computing its metrics.
Our experiments used collision-free settings with $\beta \leq 0.5$.


The simulator starts by reading a point cloud, representing each point as an FLS.  
It computes the pairwise distance between the coordinates of FLSs.
Next, it determines the FLS radius using the input $\beta$ value.  
Subsequently, the simulator adjusts the pairwise distances to reflect the distance between the surface of the FLS spheres.
This is the pose measured using a Raspberry camera and an ArUco marker.
For each FLS, the simulator computes its blind, sweet, and decaying neighbors.

The simulator constructs a graph $G=(V, E)$ where $V$ is the set of nodes, each representing an FLS, and $E$ is the set of edges that connect each FLS to its neighbors. Each edge has a weight equal to the Euclidean distance between FLSs and is labeled as blind, sweet, or decaying according to the distance between FLSs. The simulator provides interfaces to search this graph to find the shortest path between a given pair of FLSs. The interface allows optimization of the number of hops/edges in a path or the length of the path.

An experiment starts with the position of the FLSs in the ground truth.
Each FLS with a blind neighbor uses Swazure to compute its pose relative to every blind neighbor.
Subsequently, it measures the error in the estimated pose with the ground truth.
(See next section.)

The shapes considered in this paper have the same FLS density, defined as the number of FLSs per unit of surface area~\cite{swarical2024}. The mean distance between FLSs for all shapes is between 0.6 to 0.7 cm. 

\noindent{\em Quantifying error:}  
With multimedia shapes, a point cloud provides the ground truth in the relative pose between a source and a target FLS.
This is a position vector per definitions of Section~\ref{sec:term}.
We compare the magnitude of this vector with the magnitude of Swazure's estimated vector to quantify the error in distance.
The error in angle may be computed using the dot (scalar) product\footnote{$\relpose{1}{2}.\relposegt{1}{2}=||\relpose{1}{2}||_2.||\relposegt{1}{2}||_2.cos(\alpha)$ and solve for $\alpha$.} of the two vectors.
To compute the error in $\theta$ and $\phi$, we convert the estimated truth vector and its corresponding ground truth vector into their spherical coordinates; see the first paragraph of Section~\ref{sec:term}.
The difference between their $\theta$ and $\phi$ is the observed error.

\subsection{Experimental Results}
This section evaluates Swazure 
1) using Fewest-Hops versus Shortest-Length, 
and
2) moving an obstructing FLS versus a source FLS.

\begin{figure}[htbp]
    \centering
    \begin{subfigure}{0.49\columnwidth}
        \centering
        \includegraphics[width=\textwidth]{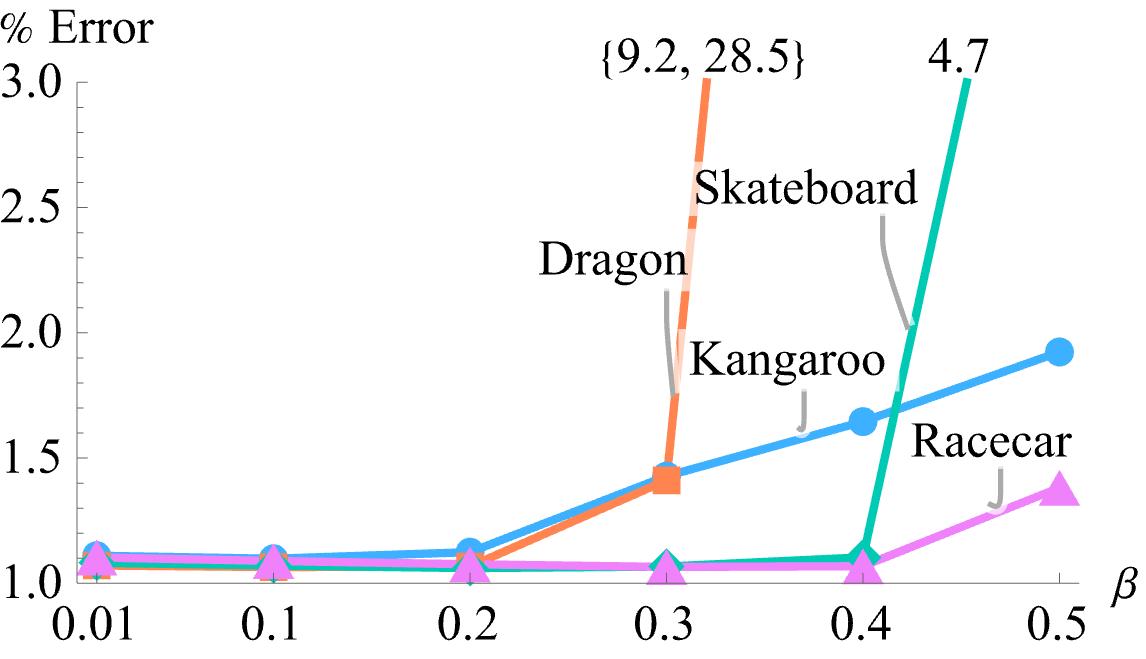}
        \caption{Fewest-Hops.}
        \label{fig:avg-dist-err-hops}
    \end{subfigure}
    \begin{subfigure}{0.49\columnwidth}
        \centering
        \includegraphics[width=\textwidth]{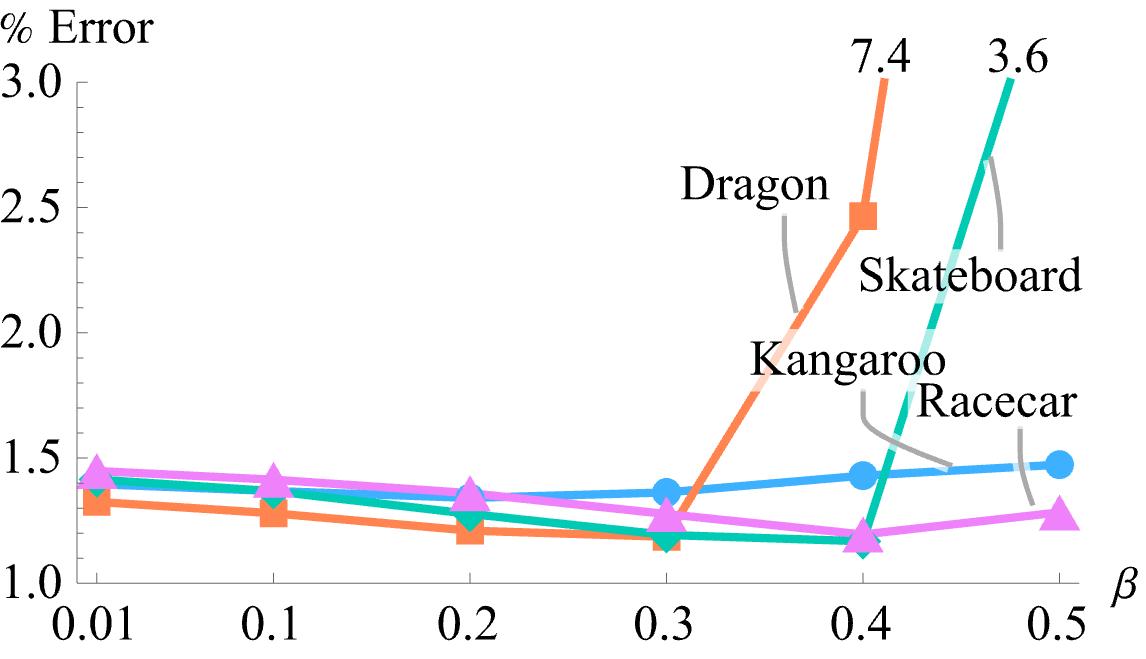}
        \caption{Shortest-Length.}
        \label{fig:avg-dist-err-length}
    \end{subfigure}
    \caption{Average percentage error in estimated distance ($\Delta$).}
    \label{fig:avg-dist-err}
\end{figure}

\subsubsection{Fewest-Hops versus Shortest-Length}\label{sec:evalpaths}

Figure~\ref{fig:avg-dist-err} shows the {\em average} percentage error with Fewest-Hops and Shortest-Length as a function of $\beta$.
Note that the scale of the y-axis is lower than 3\% with both techniques and all shapes with $\beta \leq 0.3$.
Both incur the highest error with the Dragon and $\beta$=0.5.
This is because Swazure uses decaying paths extensively with the Dragon; see the last column of Table~\ref{tab:nopath}.
While Shortest-Length has a percentage error of 7.5\%, Fewest-Hops has a percentage error of 28.5\%.
We attribute the superiority of Shortest-Length to the geometry of the Dragon.

\begin{figure}
    \centering
    \includegraphics[width=\linewidth]{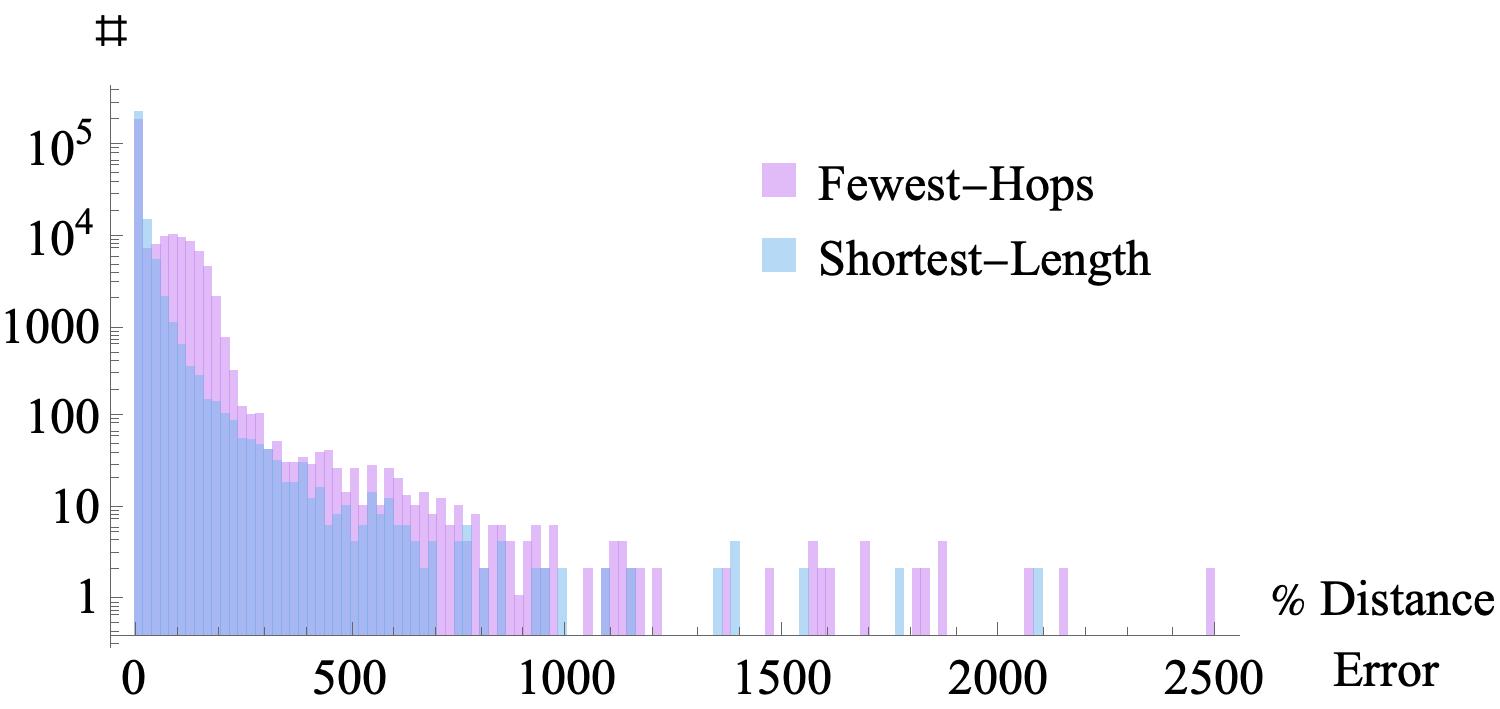}
    \caption{
    Distribution of observed percentage error in distance measurement.
    }
    \label{fig:dragon-dist-err}
\end{figure}

\begin{figure}[htbp]
    \centering
    \begin{subfigure}{0.49\columnwidth}
        \centering
        \includegraphics[width=\textwidth]{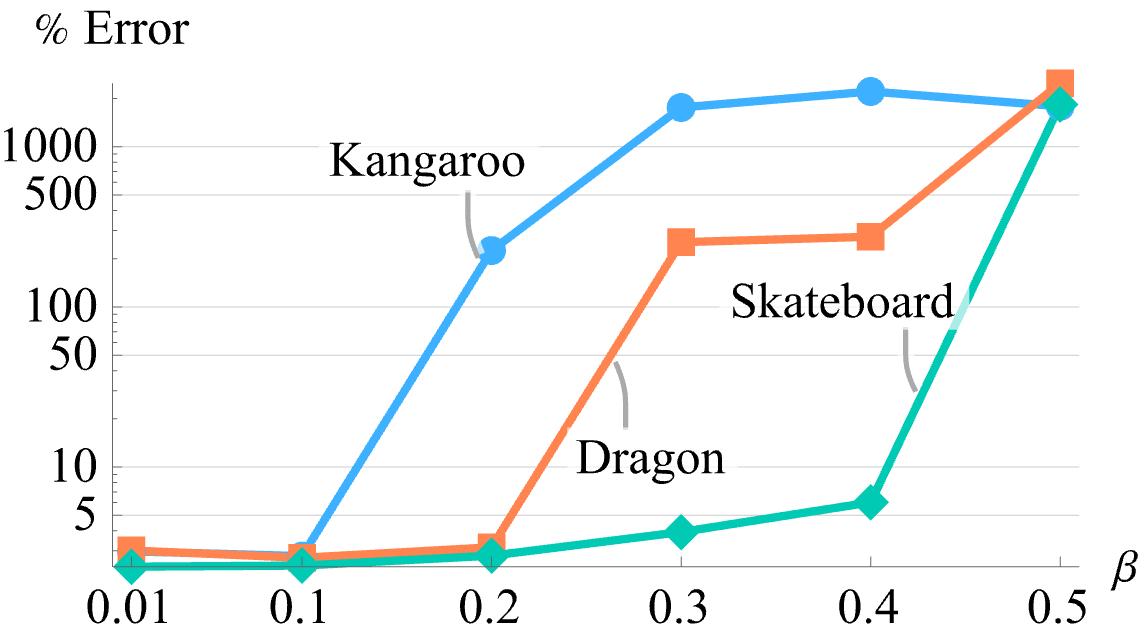}
        \caption{Fewest-Hops.}
        \label{fig:max-dist-err-length-hops}
    \end{subfigure}
    \begin{subfigure}{0.49\columnwidth}
        \centering
        \includegraphics[width=\textwidth]{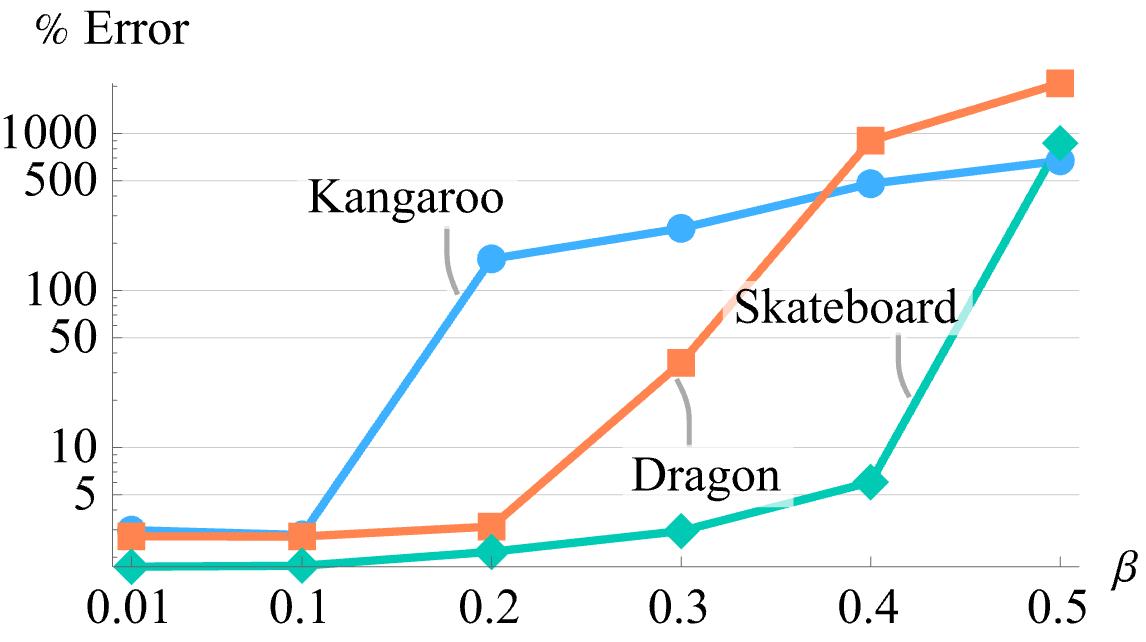}
        \caption{Shortest-Length.}
        \label{fig:max-dist-err-length-length}
    \end{subfigure}
    \caption{Maximum percentage error in estimated distance ($\Delta$).}
    \label{fig:max-dist-err-length}
\end{figure}

Figure~\ref{fig:max-dist-err-length} shows the {\em maximum} percentage error in the distance with three different\footnote{The other shapes are similar to the Skateboard with the knee of the curve ($\beta$=0.3 and 0.4) close to zero except for $\beta$=0.5.} shapes.
The y-axis is log-scaled.
In general, Shortest-Length provides a lower maximum error. 
The trends with the two techniques are similar due to their use of decaying hops. 
The x-axis of Figure~\ref{fig:dragon-dist-err} is the error of all considered paths with the Dragon.
Its y-axis is the number of times each error is incurred.
Both Fewest-Hops and Shortest-Length exhibit a long tail phenomenon by using hops with a high percentage error.
These explain the high {\em maximum} percentage error in Figure~\ref{fig:max-dist-err-length}.

\begin{figure}[htbp]
    \centering
    \begin{subfigure}{0.49\columnwidth}
        \centering
        \includegraphics[width=\textwidth]{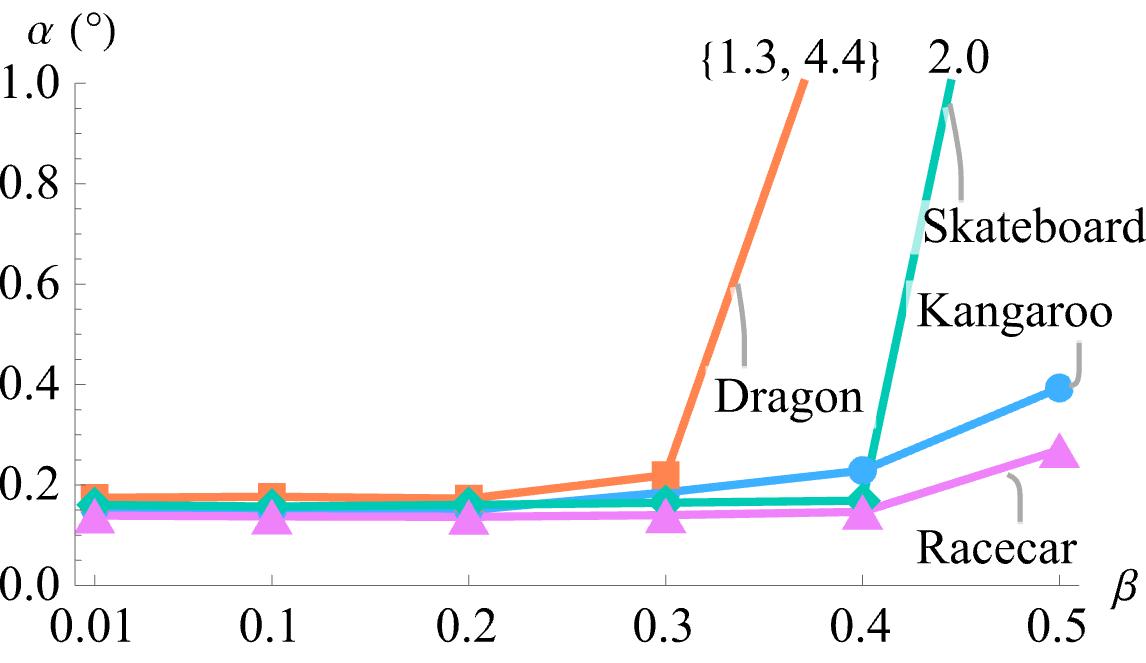}
        \caption{Fewest-Hops.}
        \label{fig:avg-alpha-hops}
    \end{subfigure}
    \begin{subfigure}{0.49\columnwidth}
        \centering
        \includegraphics[width=\textwidth]{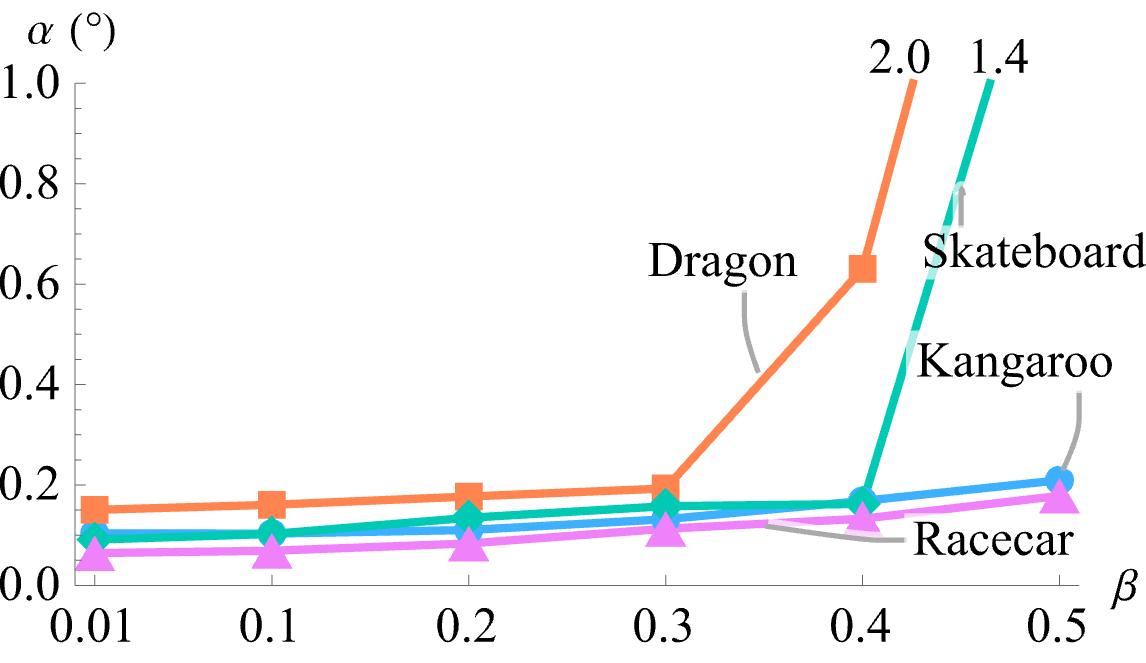}
        \caption{Shortest-Length.}
        \label{fig:avg-alpha-length}
    \end{subfigure}
    \caption{Average $\alpha$ between ground truth pose and estimated pose.}
    \label{fig:avg-alpha}
\end{figure}

With the angles $\alpha$, $\theta$, and $\phi$, the maximum error is 180 degrees.
The {\em average} error in these angles is less than 1 degree with $\beta \leq 0.4$ and all shapes except the Dragon and the Skateboard, see Figure~\ref{fig:avg-alpha} for $\alpha$.
With the two shapes, Shortest-Length provides more accurate angles with $\beta=0.5$.


\subsubsection{Move Obstructing and Move Source}\label{sec:evalmove}

These techniques enhance the Shortest-Length and Fewest-Hops in two ways:
1) by converting decaying hops into sweet hops to reduce error, and 2) by enabling blind pairs that could not estimate their pose due to an obstructing FLS.
Both techniques provide the highest benefit with $\beta$=0.5.
All shapes use a decaying path with this $\beta$ value, see last column of Table~\ref{tab:nopath}.
Both techniques minimize this use without eliminating it altogether; see the last column of Tables~\ref{tab:nopathMO}-\ref{tab:nopathMS10}.
Hence, typically the {\em average} percentage error in distance is reduced while the {\em maximum} error observes little to no improvement.
The exact amount of reduction depends on the geometry of a shape.  
For example, with the Kangaroo, the percentage improvement in accuracy ranges from 0.7\% with Move Source and $T$=2 to 8.5\% with Move Obstructing.
With a challenging shape like the Dragon, the percentage improvement is 8.2\% and 19.2\%, respectively.
Even though the percentage improvements are significant, the change in absolute values is small.
For example, with the Dragon and its 19.2\% improvement, the percentage error improves from 7.5\% to 6\%.
We speculate the significance of this is application specific.

The observations with $\alpha$, $\theta$, and $\phi$ are similar to those with distance:
A significant percentage change in the {\em average} error with little to no change in the {\em maximum} error.
The absolute change in the {\em average} angle error is typically less than 1$^\circ$.

\section{Related Work}\label{sec:related}
To the best of our knowledge, Swazure is novel and has not been described in the literature.

{\em Physical data independence} is an important concept in the area of database management systems~\cite{date1971}.  It means the organization of data is separate and independent of a physical storage device, i.e., tape, disk, DRAM, etc.
We are adopting this concept for use with FLSs and 3D displays.
Swazure is a novel algorithm that measures relative poses independently of the physical sensor.
It enables devices to cooperate to compute their relative pose at ranges that are below the minimums supported by a physical sensor.

Swazure is a building block of a localization technique
such as SwarMer~\cite{alimohammadzadeh2023swarmer} or Swarical~\cite{swarical2024} for illuminating point clouds or those for robot self exploration and mapping (SLAM)~\cite{antslam2010,kegeleirs2021swarm,zhang2024,10582478}.
A localization technique typically addresses challenging topics such as drone failure~\cite{reliability2024}, scalability, and a global reference frame, among others.
It relies on
inter-FLS position measurements. These position measurements are provided by Swazure.
Swazure is a simple building block that enables a robot to quantify its pose relative to another robot or an identified landmark.
We are not aware of a localization technique that (a) classifies the range of a sensor as blind, sweet, and decaying or (b) uses the sweet range of a sensor to quantify a pose for a blind range.

The use of communication for pose estimation in drones is well-established, as demonstrated in studies such as~\cite{7440990,8442959,10606514,10449450,4671092,8360450,9363562}.
The most relevant is~\cite{7440990}, which utilizes fiducial markers and cameras mounted on UAVs.
This study identifies a blind range for the camera and operates under the assumption of two drones. When one drone becomes blind, the other tracks it and provides the relative pose information.
While it does not consider more than 2 drones, its extensions to 3 or more drones 
may use Swazure to provide the relative pose of multiple drones.
Swazure also uses communication between FLSs.
It is different and novel in several ways.
First, it abstracts the range of a sensor into
blind, sweet, and decaying, using a sweet path to bypass blind and decaying ranges to enhance the accuracy of estimated poses.
These paths consist of at least one intermediary FLS, requiring the participation of a minimum of 3 FLSs.
The number of FLSs in a path depends on the geometry of a shape and the value of $\beta$.
For example, with the Dragon and $\beta$=0.5, an average of 4.3 FLSs and a maximum of 14 FLSs constitute a sweet path.
Second, Swazure includes two techniques, Move Obstructing and Move Source, to facilitate line of sight in the presence of obstructions.
\vspace{-0.1in}
\section{Conclusions}\label{sec:conc}
Swazure provides for the physical independence of an FLS from the operating range of its sensor.
It enables the FLS to compute its relative pose to another FLS at ranges not supported by its sensor.
It realizes this by abstracting the range of a sensor into {\em blind} (range without the ability to quantify pose), {\em sweet} (most accurate range), and {\em decaying} (range with an increased noise as a function of some parameter such as distance).
Two FLSs in a blind range compute their relative pose with a high accuracy by computing a path of intermediary FLSs that are sweet neighbors.
When no sweet paths are available, Swazure may use a path that involves decaying neighbors or a combination of sweet and decaying neighbors.
We introduced Fewest-Hops and Shortest-Length as a technique to select between multiple candidate paths. 
Experimental results show the superiority of Shortest-Length.  
We identified the size of an FLS relative to the minimum required distance between FLSs as an important parameter.
Their ratio, $\beta$, impacts the number of neighboring FLSs in a blind range with no sweet paths due to obstruction. 
We introduced two techniques, Move Obstructing FLS and Move Source FLS, to address this challenge.
An evaluation of these techniques shows the superiority of Move Obstructing.

\balance

\begin{acks}
This research was supported in part by the NSF grants IIS-2232382 and CMMI-2425754.
We gratefully acknowledge  CloudLab~\cite{emulab} for the use of their resources to enable all experimental results presented in this paper.
\end{acks}


\bibliographystyle{ACM-Reference-Format}
\bibliography{refs}

\end{document}